\documentclass[conference]{IEEEtran}

\pdfoutput=1

\usepackage{amsmath}
\usepackage{amsfonts}
\usepackage{graphics, graphicx}
\usepackage{cite}
\usepackage{multirow}
\usepackage{url}
\usepackage{enumerate}
\usepackage{etoolbox}
\usepackage[ruled,norelsize]{algorithm2e}
\usepackage{numprint}
\newcommand{\subparagraph}{}
\usepackage{titlesec}
\usepackage{todonotes}

\newcommand{\norm}[1]{\left\lVert#1\right\rVert}

\npthousandsep{,}\npthousandthpartsep{}\npdecimalsign{.}

\titlespacing{\section}{0pt}{4pt}{4pt}
\titlespacing{\subsection}{0pt}{3pt}{3pt}

\makeatletter
\newcommand{\removelatexerror}{\let\@latex@error\@gobble}
\makeatother

\DeclareMathOperator*{\argmin}{argmin}

\newtheorem{theorem}{Proposition}

\IEEEoverridecommandlockouts

\title{Deep Learning-Based Quantization of L-Values for Gray-Coded Modulation\thanks{This work has been accepted for presentation at Globecom 2019. Supported by grants ONR N00014-19-1-2590, ARO W911NF-18-1-0247, NSF CNS 1731384, and a gift from NXP Semiconductors Inc., Austin, Texas.}}

\author{\IEEEauthorblockN{Marius Arvinte, Sriram Vishwanath, and Ahmed H. Tewfik}
	\IEEEauthorblockA{Department of Electrical and Computer Engineering\\
		University of Texas at Austin\\
		Austin, Texas 78712\\
		Email: arvinte@utexas.edu}}

\makeatletter
\patchcmd{\@maketitle}
{\addvspace{0.8\baselineskip}\egroup}
{\addvspace{-0.4\baselineskip}\egroup}
{}
{}
\makeatother

\begin{document}
	
	\maketitle
	
	\begin{abstract}
		In this work, a deep learning-based quantization scheme for log-likelihood ratio (L-value) storage in general fading scenarios affected by interference is introduced. We analyze the dependency between the average magnitudes of different L-values and show they follow a consistent ordering, regardless of the channel coefficient or interference distribution. Based on this we design a deep autoencoder that jointly compresses and separately reconstructs each L-value, allowing the use of a weighted loss function that aims to more accurately reconstruct low magnitude inputs. Our method is shown to be competitive with state-of-the-art maximum mutual information quantization schemes, reducing the required memory footprint by a ratio of up to two and achieving a loss of performance smaller than $0.1$ dB with less than two effective bits per L-value or smaller than $0.04$ dB with $2.25$ effective bits. We experimentally show that our proposed method is a universal compression scheme in the sense that after training on an LDPC-coded Rayleigh fading scenario we can reuse the same network without further training on other channel models and codes while preserving the same performance benefits.
	\end{abstract}
	
	\section{Introduction}
	\label{section_intro}
	
	Deep learning has recently gained a foothold in wireless communications and signal processing, with experimental results showing that it can be used for various tasks ranging from channel code design \cite{kim2018deepcode} to black-box communication schemes \cite{o2017introduction}. At the same time, quantization of information in communication systems is critical for applications such as feedback, relay, or hydrid automatic repeat request schemes, which require long-term storage of information. Motivated by finding a quantization scheme that is as greedy as possible with minimal impact to the end-to-end performance, we introduce a deep learning-based L-value vector quantization scheme that leverages a statistical ordering of the average magnitudes of different bit positions to weigh its loss function during training. 
	
	Log-likelihood ratio quantization is a well studied topic, where it has been identified that the optimal formulation aims to maximize mutual information \cite{rave2009quantization}. Prior work in \cite{winkelbauer2015quantization} presents a data-driven approach for quantizing L-values even in cases when the channel has an arbitrary or intractable distribution. The work in \cite{danieli2010maximum} introduces a maximum mutual information vector quantization method for L-values by using the Lloyd algorithm with a KL divergence metric. In both cases, a sharp drop in performance is exhibited when the effective storage size approaches or goes below two bits per L-value.
	
	More recently, deep networks have been used for quantization and codebook learning, where the main challenge is back-propagating gradients through the quantization function. Solutions to this include soft-to-hard approximations of the nearest neighbor function that are parameterized by an increasing attraction coefficient over time or replacing the null gradient with other approximations. The work in \cite{agustsson2017soft} uses such a schedule during the training phase to learn a compression scheme for high resolution images that outperforms state-of-the-art schemes by almost an order of magnitude. With the same goal, the work in \cite{van2017neural} introduces the architecture of vector quantized variational autoencoder (VQ-VAE) to learn a discrete, compressed representation of image and speech signals by approximating the null gradient with a linear function.
	
	Even though these solutions are shown to be successful for images, in the case of communications the signal statistics are fundamentally different and any application requires careful analysis and design of the network architecture and loss function. Previous work in \cite{lu2018mimo} uses an autoencoder structure to compress the channel state information (CSI) matrix by exploiting its time and spatial domain correlation. In this work, we propose to exploit the use of binary reflected Gray coding (BRGC), almost universally adopted in practical systems and known to be optimal \cite{agrell2006gray} and show that it induces an ordering on the average magnitude of the L-values that holds for any channel coefficient distribution. We leverage this to design a deep autoencoder network that uses a branched decoder architecture to individually reconstruct each L-value, allowing us to weigh the loss function towards smaller magnitude L-values.
	
	Our work is closest to \cite{DBLP:journals/corr/abs-1903-04656}, where a deep autoencoder is used to jointly compress and reconstruct the set of L-values corresponding to a channel use by leveraging the fact that three sufficient statistics will recover them. However, their work exhibits a performance gap even when the compressed signal (latent representation) is not quantized. By leveraging the statistics of L-values, we manage to virtually close this performance gap, reducing it from $0.2$ dB to as low as $0.02$ dB and consequently improving the quantized results as well. We compare our work with the results in \cite{DBLP:journals/corr/abs-1903-04656} and state-of-the-art maximum mutual information schemes and show that we exhibit a compression gain of up to two, allowing us to use an effective storage size of less than two bits per L-value for high order modulation schemes. Furthermore, we experimentally show that the same network weights have generalization properties and can be used for a wide range of scenarios with different channel models and codes, without requiring any further training or adjustments.
	
	\section{System Model}
	\label{section_sys}

	Consider the digital baseband model of a binary reflected Gray coded (BRGC) $M$-QAM modulation scheme, where the transmitted symbol $x$ is obtained by mapping the bits $b_1, \dots, b_K$ to a complex symbol belonging to the constellation $C$, where $K = \log_2 M$. We assume there are a number of $N_I$ interfering symbols denoted by $z_i$, each independently drawn from a constellation $C_I$. Under a flat fading complex channel model the complex received symbol $y$ has the expression
	\begin{equation}
		y = hx + \sum_{i=1}^{N_I} h^{(I)}_i z_i + n = hx + \langle \mathbf{h}^{(I)}, \mathbf{z} \rangle + n,
	\end{equation}

	\noindent where $n$ is drawn from a circular complex Gaussian distribution with zero mean and variance equal to $\sigma_n^2$ and $h$ and $h^{(I)}_i$ represent the complex channel coefficients of the desired signal and each interferer, respectively. Given complete channel state information, the exact log-likelihood ratio (L-value) at the receiver for bit $b_k$ is given by
	\begin{equation}
	    L_k = \log{ \frac{P(y|b_k = 1)}{P(y|b_k=0)} } = \log{\frac{\sum\limits_{\mathbf{z} \in C_I} P(y|b_k=1, \mathbf{z}) P(\mathbf{z})}
	    {\sum\limits_{\mathbf{z} \in C_I} P(y|b_k=0, \mathbf{z}) P(\mathbf{z})}}.
	\end{equation}
	
	Assuming that the prior probability of the interference vector $\mathbf{z}$ is uniform, we can simplify the expressions and expand each conditional probability as
	\begin{equation}
	    L_k = \log{\frac{\sum\limits_{\mathbf{z} \in C_I} \sum\limits_{\hat{x} \in C, b_k=1} \exp{ -\frac{|y-h\hat{x} - \langle \mathbf{h}^{(I)}, \mathbf{z} \rangle|^2}{\sigma_n^2}}}
	    {\sum\limits_{\mathbf{z} \in C_I} \sum\limits_{\hat{x} \in C, b_k=0} \exp{ -\frac{|y-h\hat{x} - \langle \mathbf{h}^{(I)}, \mathbf{z} \rangle|^2}{\sigma_n^2}}}}.
	\end{equation}
	
	Carrying an analysis similar to that in \cite{DBLP:journals/corr/abs-1903-04656} and factoring out the $h$ term from the exponents we derive the sufficient statistics required to exactly reconstruct the set of L-values as
	\begin{equation}
	    \label{eq_stats}
	    G = \left( \frac{\lvert h \rvert}{\sigma_n} \right) ^2, \tilde{y} = \frac{y}{h} \ \text{and} \ \tilde{z}_i = \frac{h^{(I)}_i}{h}.
	\end{equation}

	Counting each complex value as two real ones, it follows that the number of real-valued sufficient statistics is equal to $3 + 2 \times N_I$ for a scenario with $N_I$ interferers. The special case $N_I = 0$ corresponds to the case where there is no interference, or, more interestingly, when there is no information about the interference channel gains $\mathbf{h}$. In this case, the unknown interference is effectively treated as noise.

	We further consider the conditional distributions of the L-values under the max-log approximation for different bit positions $k = 1, \dots, K$. The work in \cite{alvarado2009distribution} shows that in a nonfading (i.e., affected only by noise) scenario the probability density function (PDF) of the L-values for each bit level can be written as a sum of truncated Gaussian PDFs. According to \cite{alvarado2009distribution}, the approximate PDF of the $k$-th L-value conditioned on the transmitted bit $b_k$ has the expression
	\begin{equation}
		p_k(\lambda | b_k) = \sum\limits_{l = 0}^{\frac{M}{2}-1} w_{k, l} \Phi(\lambda; (-1)^{b_k+1} \mu_l, \sigma_l^2),
		\label{eq_awgn_pdf}
	\end{equation}
	
	\noindent where $\Phi(\lambda; \mu, \sigma^2)$ represents the Gaussian PDF and $w_{k,l}$ is the uniform probability that $L_k$ is a Gaussian with mean and variance given by $\mu_l$ and $\sigma_l^2$ respectively, in concordance with the Zero-Crossing Model approximation in \cite{alvarado2009distribution}. As the authors of \cite{alvarado2009distribution} show, this leads to a key observation about the L-values corresponding to different bit positions, namely that after splitting the bits in two groups (due to the real/imaginary symmetry induced by BRGC) the bits occupying the first positions are more robust to channel conditions than later ones. Formally, this can be expressed by the inequality
	\begin{equation}
		\label{eq_ordering}
		\mathbb{E} \left[ |L_1| \right] > \mathbb{E} \left[ |L_2| \right] > \dots > \mathbb{E} \left[ |L_{\frac{K}{2}}| \right],
	\end{equation}
	
	\noindent where the expectation is taken across the noise and a similar ordering holding true for the second half of L-values associated with the imaginary part. Finally, let $\Lambda_k$ be the soft bit associated to the L-value $L_k$, given by \cite{rave2009quantization}
	
	\begin{equation}
		\Lambda_k = \tanh{\frac{L_k}{2}}.
	\end{equation}
	
	Since $\tanh$ is a monotonic, increasing function it follows that the ordering in \eqref{eq_ordering} holds for $\Lambda_k$ as well. We let $\mathbf{L}$ and $\mathbf{\Lambda}$ denote the $K$-dimensional real, ordered -- according to \eqref{eq_ordering} -- vectors of L-values and soft bits corresponding to a single channel use with $\Lambda_k$ and $L_k$ their $k$-th elements, respectively. Our goal is to compress $\mathbf{\Lambda}$ to a three-dimensional latent representation, quantize it to a finite, small number of bits, and reconstruct the original input.

	The basic machine learning structure we use for compressing the L-values is a deep neural network. From a high-level perspective, a deep neural network can be viewed as a parameterized function $f(\mathbf{x}; \theta_f)$, where $\mathbf{x}$ is the real-valued input vector and $\theta_f$ denotes the weight vector, containing the serialized weights of all layers. For the rest of this work, we only refer to feedforward, fully-connected neural networks, in which a layer that takes as input the vector $\mathbf{x}$ and outputs $\mathbf{y}$ implements the operation
	
	\begin{equation}
		\mathbf{y} = \phi(\mathbf{W}\mathbf{x} + \mathbf{b}),
	\end{equation}
	
	\noindent where $(\mathbf{W}, \mathbf{b})$ represent the weights and biases associated with a layer and $\phi$ is the element-wise activation function and all dimensions are consistent with matrix-vector multiplication. Typical activation functions include the rectified linear unit (ReLU) given by $\phi(x) = \max{\{x, 0\}}$ and the hyperbolic tangent $\phi(x) = \tanh x$. Finally, an autoencoder is a deep neural network that is trained to reconstruct its own input $\mathbf{x}$. This is commonly achieved by performing gradient updates on the weights $\theta_f$ in order to minimize the empirical risk function
	
	\begin{equation}
		\mathcal{L}(\theta_f) = \frac{1}{N} \sum\limits_{i=1}^{N} L(f(\mathbf{x}_i; \theta_f), \mathbf{x}_i),
	\end{equation}
	
	\noindent where $\mathbf{x}_i$ represents the $i$-th training sample and $L$ can be chosen to any distance or quasi-distance function. Importantly, we note that gradient-based approaches are incompatible with the quantization of hidden activations in deep networks, since the gradient becomes null almost everywhere after quantization, preventing preceding weights from being updated.
	
	\section{The Proposed Scheme}
	\label{section_prop_alg}
	
	While the original derivation is performed for a nonfading scenario, additional conditioning on the channel realization and averaging ensures the ordering of L-values holds even for an arbitrary fading distribution. We formulate and prove the following proposition.
	
	\begin{theorem}
		The ordering in \eqref{eq_ordering} holds for any arbitrary distribution of $h$.
		\label{prop_1}
	\end{theorem}

	\begin{IEEEproof}
		We prove this for the interference-free case, but the proof can be extended for interference, by taking the double integral into account. First we note that by \eqref{eq_stats} knowledge of $|h|^2$ is sufficient to characterize the distribution of L-values, since the phase can always be corrected assuming full CSI is available. By letting $g = |h|^2$ and expanding the PDF of the $k$-th L-value we obtain
		\begin{equation}
			p_k(\lambda | b_k) = \int_{0}^{\infty} p_k(\lambda | b_k, g) f_g(g) \ dg,
		\end{equation}
		\noindent where $f_g$ is the PDF of $g$. Considering $p_k(\lambda | b_k, g)$, it follows that \eqref{eq_awgn_pdf} holds for any fixed $g$, thus the ordering \eqref{eq_ordering} holds pointwise, thus it holds for any $f_g(g)$.
	\end{IEEEproof}
	
\begin{figure}[!t]
		\centering
		\includegraphics[width=3.4in]{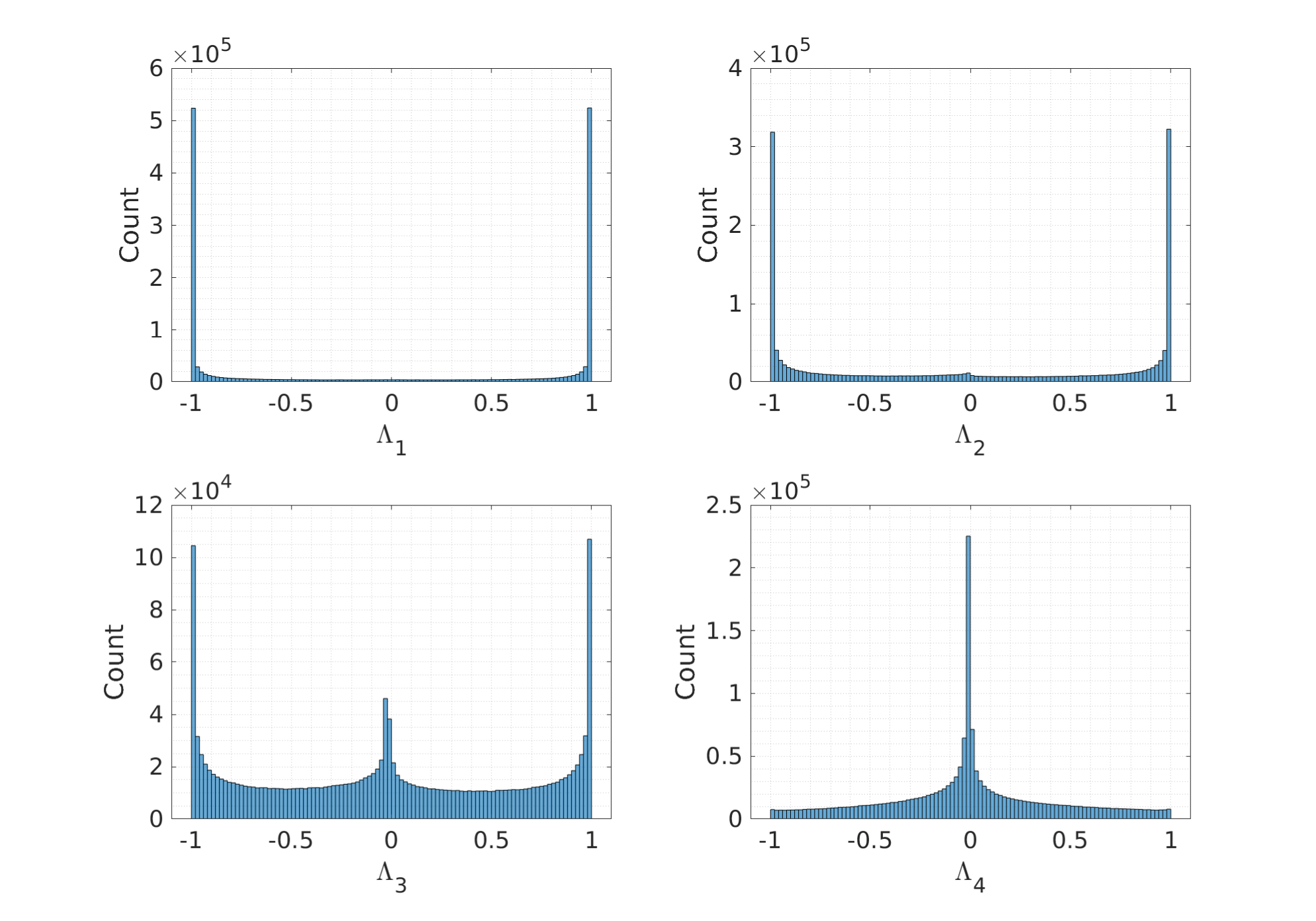}
		\caption{Empirical PDFs for the first four soft bits of a $K=8$ ($256$-QAM) transmission over an interference-free Rayleigh fading channel at $18$ dB, exhibiting the ordering in \eqref{eq_ordering}. The next four soft bits have a similar order due to the binary reflected Gray coding.}
		\label{fig:example_histograms}
	\end{figure}

	The previous result is proven for the max-log approximation of the L-values but empirically holds for the complete expression as well. In the case of an interference-free Rayleigh fading channel, \figurename{ \ref{fig:example_histograms}} illustrates this property by plotting four of the eight distributions of the L-values for a 256-QAM scenario, where it can be observed that the latter bits have a lower average absolute reliability than the earlier ones. This asymmetry of the different bit locations affects the performance of forward error correction, especially in the mid-high signal-to-noise ratio regime, where accurate reconstruction of low magnitude L-values is shown to be critical for correct decoding \cite{zhang2013quantized}.
	
\begin{figure}[t]
		\centering
		\includegraphics[width=3.5in]{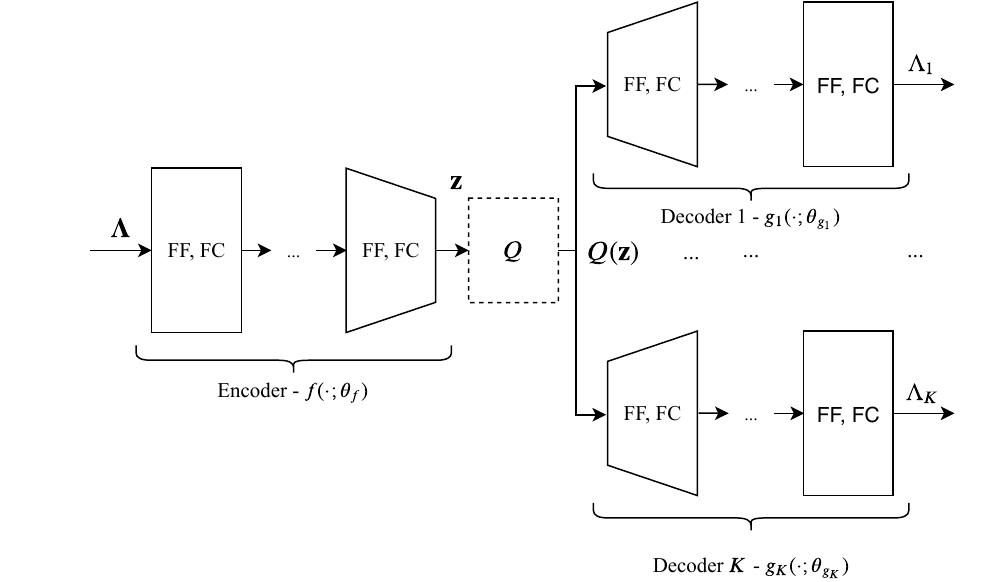}
		\caption{Block diagram of our proposed deep network architecture with one decoder for each soft bit. Each solid rectangular box represents a feedforward, fully-connected layer. The dashed block represents the latent representation quantizer and is not active during training.}
		\label{fig:block_arch}
	\end{figure}

	Equations \eqref{eq_stats} and \eqref{eq_ordering} motivate the architecture of our compression and reconstruction network. We use an autoencoder with a latent representation of dimension $3 + 2 \times N_I$, a joint encoder and a branched decoder (i.e., one, smaller, deep neural network is used to separately reconstruct each soft bit). The architecture of our solution is shown in \figurename{ \ref{fig:block_arch}}. The network takes as input the vector of soft bits $\mathbf{\Lambda}$ and feeds them to the encoder, where the compressed latent representation $\mathbf{z}$ is output. Letting $f(\cdot; \theta_f)$ be the encoder part of the network and $g_k(\cdot; \theta_{g_k})$ each of the bit decoders, the $k$-th reconstructed soft bit can be expressed as
	
	\begin{equation}
		\tilde{\Lambda}_k = g_i \left( \mathbf{z} ; \theta_{g_k} \right) =  g_i\left(f\left( \mathbf{\Lambda} ; \theta_f \right) ; \theta_{g_k} \right).
	\end{equation}
	
    To account for accurate reconstruction of low magnitude soft bits, we adopt two measures:
	\begin{enumerate}
		\item We use a sample-wise weighted mean squared error function as in \cite{DBLP:journals/corr/abs-1903-04656}. The loss function between the $k$-th soft bit and its reconstruction has the expression
		
		\begin{equation}
			L (\tilde{\Lambda}_k, \Lambda_k) = \frac{|\tilde{\Lambda}_k - \Lambda_k|^2}{|\Lambda_k| + \epsilon},
		\end{equation}
		\noindent where $\epsilon = 10^{-6}$ is used for numerical stability. This formulation ensures that more importance is given to soft bits with low values inside the same bit position.
		
		\item We use a set of real weights $w_k \ge 0$ to weigh the contributions of each soft bit to the total loss function leading to the expression
		
		\begin{equation}
			\mathcal{L(\tilde{\mathbf{\Lambda}}, \mathbf{\Lambda})} = \sum\limits_{k=1}^K w_k \sum\limits_{i=1}^{N} L(\tilde{\Lambda}_k^{(i)}, \Lambda_k^{(i)}),
		\label{eq_loss}
		\end{equation}
		\noindent where the weights are normalized to satisfy $\sum_k w_k = 1$, thus at least one weight must be strictly positive.
	\end{enumerate}

	Note that the weights $w_k$ are not applied per sample, but rather per soft bit and ensure that more importance is placed on reconstructing certain bit positions. By using Proposition \ref{prop_1} and the ordering in \eqref{eq_ordering}, we order the weights $w_k$ in decreasing order of reliability by
	
	\begin{equation}
		\begin{split}
			w_1 \le w_2 \le \dots \le w_{\frac{K}{2}}, \\
			w_{\frac{K}{2} + 1} \le w_{\frac{K}{2} + 2} \le \dots \le w_K.
		\end{split}
	\label{eq_order}
	\end{equation}
	
	Once the weights are set, the architecture is jointly trained for a number of epochs. Since we are using a single encoder, all gradient updates are averaged (with the weights $w_k$ factored in) when $\theta_f$ is updated, while $\theta_{g_k}$ are individually updated for each of the constituent decoders. We note that a different training regime is also possible where only the weights of specific decoders are updated if the performance after joint training (as measured by the loss function applied component-wise) is not good enough on specific soft bits. In fact, the scheme is completely modular in terms of the decoders, meaning that we can replace any of them with other reconstruction methods once the encoder is fixed.
	
	Once training is complete, we obtain a universal, compressed representation of the soft bits in the form of $\mathbf{z} = f(\mathbf{\Lambda})$, which needs to be quantized, stored and reconstructed during inference. Letting $\hat{\mathbf{z}}_i, i = 1, \dots, 2^{N_b}$ be the $N_b$-bit codebook in the latent space the quantization function is given by

	\begin{equation}
		Q(\mathbf{z}) = \argmin_{\hat{\mathbf{z}}} \norm{\mathbf{z} - \hat{\mathbf{z}}}_2.
	\end{equation}
	
	The use of a minimum distortion quantization in the latent space is justified by prior work showing that over-parameterized deep neural networks are naturally robust when their hidden activations are quantized \cite{guo2018survey}. Finally, reconstruction of the original L-value vector is performed by applying the decoders for each soft bit on the quantized latent representation with $\hat{\Lambda}_k = g_k(\hat{\mathbf{z}})$.
	
	\section{Performance Results}
	\label{section_perf}
	\subsection{Architectural and Training Details}
	
	We use a number of three hidden layers for the encoder and each of the decoders with a universal intermediate output size of $4K$, except for the latent representation which has a dimension of three as discussed in Section \ref{section_prop_alg}. Storing weights in 32-bit floating precision leads to a total memory footprint (considering the quantization codebook as negligible) lower than $83$ KB for $K = 8$ ($256$-QAM), and scaling on the order of $K^2$, thus rapidly decreasing for lower order modulations.
	
	All hidden activations are ReLU, except for the latent representation and the outputs, which come from $\tanh$ activations. During training, we also add a small amount of additive white Gaussian noise with zero mean and $\sigma_t = 10^{-3}$ to the latent representation before it is decoded to encourage generalization and robustness to numerical quantization. Since the latent representation $\mathbf{z}$ comes from the $\tanh$ activation, each of its elements is bounded to the $[-1, 1]$ interval and the architecture learning trivial solutions such as boosting the power of the latent representation to overcome the added noise.
	
	\begin{figure}[t]
		\centering
		\includegraphics[width=2.7in]{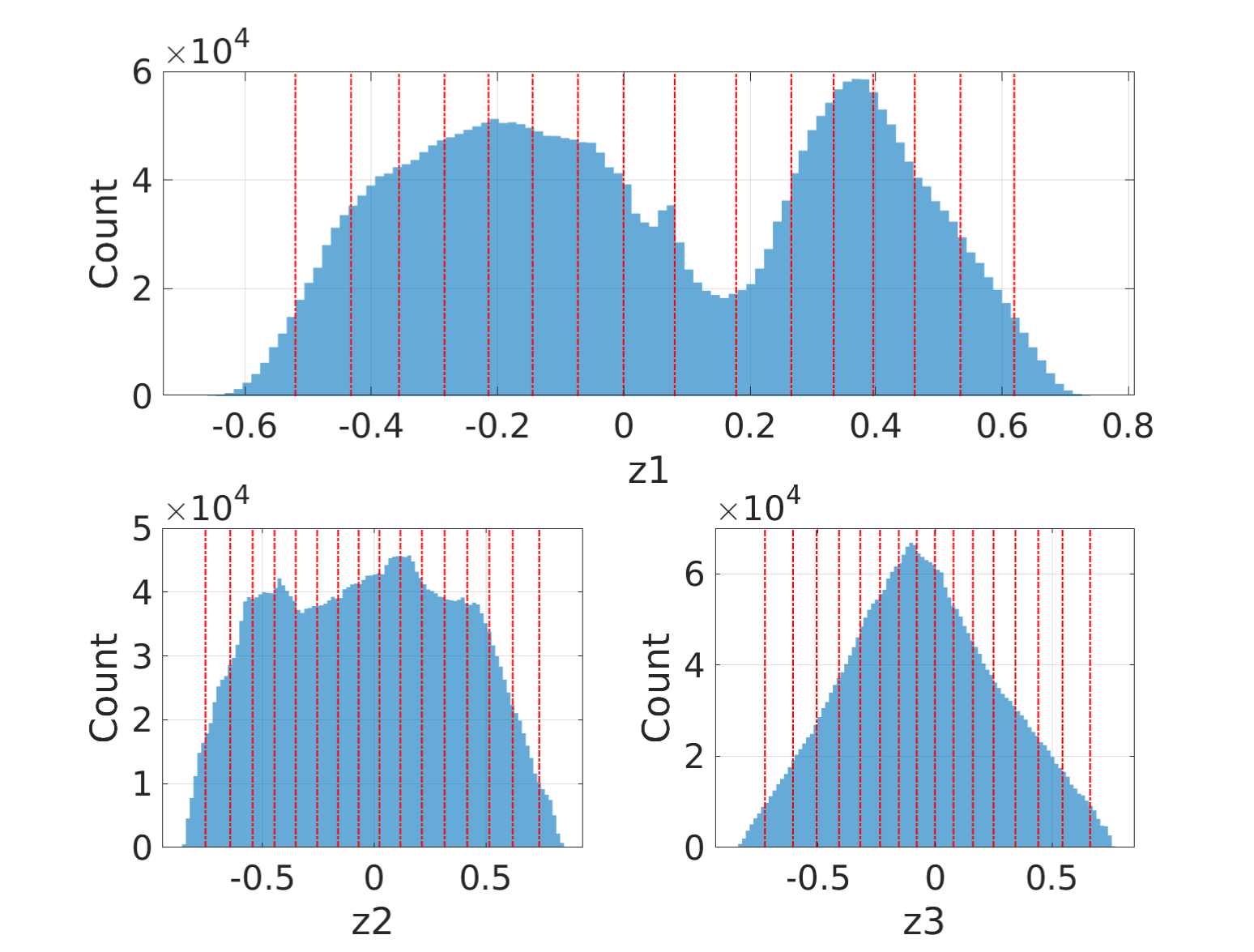}
		\caption{Histograms of the three components of $\mathbf{z}$ after training and the resulted one-dimensional k-Means quantization codebooks with $4$ bits ($16$ levels) per latent component in dashed red lines. The Cartesian product of these quantizers is used to derive the $Q$ function. This latent representation is only trained once and reused for different channel models and codes.}
		\label{fig:kmeans}
	\end{figure}
	
	We use the AMSGrad version of the Adam algorithm for minimizing the empirical risk function in \eqref{eq_loss} with a batch size of $\numprint{65536}$ samples, learning rate of $0.001$ and recommended parameters \cite{reddi2019convergence} and split the training in two phases. The entire procedure is summarized in Algorithm \ref{algo_main}. In the first phase the encoder and all decoders are jointly trained starting with equal weights $w_k = \frac{1}{K}$. After a number of epochs $N_{e_1}$ the average reconstruction error $e_k$ is computed for all soft bits in the training data and we update the weights $w_k$ as
	
	\begin{equation}
		w_k = \frac{e_k}{\sum\limits_{i=1}^K e_i}.
		\label{eq_weight_update}
	\end{equation}
	
	This process is repeated for a number of rounds $N_r$, placing a higher importance on soft bits that have larger average reconstruction errors. We empirically observe that this procedure always respects the order induced by \eqref{eq_order} and also converges to a steady state where the weights no longer need an update. Alternatively, if a closed form expression of the distribution of $p_k(\lambda|b_k)$ is available and can be numerically evaluated one can use a fixed set of weights inversely proportional to $\mathbb{E}[|L_k|]$.
	
	\begin{figure}[!t]
		\removelatexerror
		\begin{algorithm}[H]
			\caption{Training procedure}
			\label{algo_main}
			\KwData{$\mathbf{\Lambda}_i \in \mathbb{R}^K, i = 1, \dots, N.$}
			Randomly initialize weights $\theta_f, \theta_{g_k}$. \\
			Initialize weights $w_k = \frac{1}{K}, \forall k = 1, \dots, K.$ \\
			\emph{Stage 1}. Encoder and decoders jointly optimized. \\
			\For{$i = 1$ to $N_r$}
			{
				Train with fixed $w_k$ for a number of $N_{e_1}$ epochs. \\
				\For{$j = 1$ to $N_{e_1}$}
				{
					Mini-batch Adam update of weighted loss function. \\
					\For{$b = 1$ to $N_b$}
					{
						$\theta_f, \theta_{g_k} \leftarrow \text{Adam}(\mathbf{\Lambda}_i, w_k, \theta_f, \theta_{g_k})$
					} 
				}
				Compute average reconstruction error for each L-value. \\
				$e_k = \frac{1}{N} \sum\limits_{i=1}^N L(\tilde{\Lambda}_i, \Lambda_i)$ \\
				Update $w_k$ using \eqref{eq_weight_update}. \\
			}
			\emph{Stage 2}. Individual decoder training.
			
			\For{$k = 1$ to $K$}
			{
				\For{$i = 1$ to $N_{e_{2,k}}$}
				{
					Mini-batch Adam update of loss function. \\
					$\theta_{g_k} \leftarrow \text{Adam}(\mathbf{z}_i, \theta_{g_k})$
				}
			}
		\end{algorithm}
	\end{figure}
	
	In the second phase we freeze the weights of the encoder and continue individual training for each decoder for a number of $N_{e_{2,k}}$ epochs. Here we leverage the branched architecture to improve the performance of individual decoders without affecting the learned representation. We explored options where $N_{e_{2,k}}$ is equal or proportional to the stationary $w_k$, but both cases lead to roughly the same performance. Since each decoder only updates its own weights, this process is fully parallelizable among them.

	Once training is completed, we apply a mini-batch version of the k-Means algorithm \cite{sculley2010web} to independently obtain a non-uniform scalar quantizer for each of the three components of the latent representation. Note that this has the advantage of drastically minimizing the storage requirements of the codebook versus vector quantization of the latent space and is also empirically observed to offer similar or better performance. The training data consists of L-values computed using \eqref{eq_lvalue} and generated from the coded bits of a number of $\numprint{10000}$ LDPC codewords with a length of $648$ and rate $1/2$ transmitted over a Rayleigh fading channel with $h \sim \mathcal{CN}(0, 1)$. The complete source code, pretrained networks and all performance results are available online\footnote{\url{https://github.com/mariusarvinte/deep-llr-quantization}}.
	
	\subsection{Impact of Quantization on Block Error Rate}
	\label{subsect_base_results}
		
	\begin{figure}[t]
		\centering
		\includegraphics[width=3.5in]{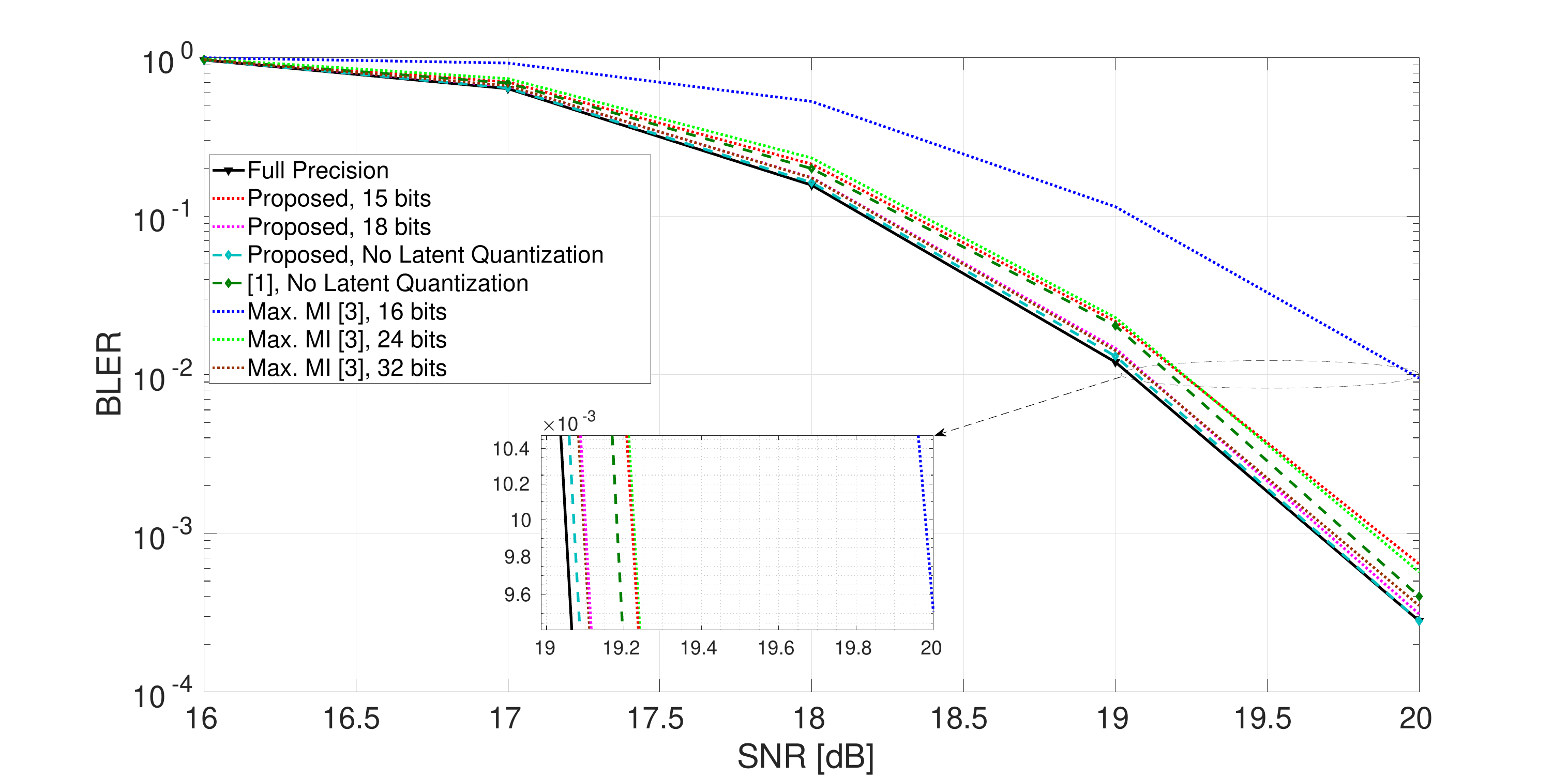}
		\caption{BLER performance of the proposed quantization scheme in a Rayleigh fading scenario with LDPC coding and $K=8$ ($256$-QAM) modulation.}
		\label{fig:ldpc_rayleigh}
	\end{figure}
	
	Throughout this section, we show the results obtained for $K = 8$ ($256$-QAM), but the architecture can be readily used for any modulation scheme. Since the number of quantized values is a constant w.r.t. $K$, the scheme offers more efficient compression for high-order modulation mappings. The first experiment involves investigating the performance of our scheme in terms of block error rate (BLER) in the same conditions in which the training data is generated. We generate $\numprint{10000}$ LDPC codewords for each signal-to-noise ratio in the set $\{16, 17, 18, 19\}$ dB, corresponding to a high-mid noise power regime, concatenate the data points and shuffle them. Once training is completed, we design the k-Means quantizers using the latent representation of the same dataset. The empirical marginal distributions of the three components of $\mathbf{z}$ are shown in \figurename{ \ref{fig:kmeans}} together with their quantization codebooks. For validation, we generate a set of unseen $\numprint{100000}$ LDPC codewords across a slightly wider signal-to-noise ratio range, encode them, quantize the latent representation with the trained codebooks and reconstruct them, followed by LDPC decoding using belief propagation with $50$ iterations.
	
	We compare the performance of our method with the maximum mutual information quantization in \cite{winkelbauer2015quantization}, where we train a reference quantization for each separate bit position to account for large order modulation schemes, as well as use an initial codebook constrained in the $[-3, 3]$ to further boost performance. Additionally, we include the full precision (unquantized latent representation) results from \cite{DBLP:journals/corr/abs-1903-04656} to show that we successfully cover the performance gap coming from the autoencoder reconstruction. \figurename{ \ref{fig:ldpc_rayleigh}} shows the obtained block error rate curves for all the schemes, as well as the unquantized performance. Comparing at BLER $=0.01$ we notice that we achieve the same performance with $15$ bits instead of $24$, leading to a compression ratio of $1.6$ times and a loss of performance smaller than $0.2$ dB when compared to full precision storage, while using an equivalent of $\frac{15}{K} = 1.875$ bits per L-value. As a contrast, using the scheme in \cite{winkelbauer2015quantization} with $2$ bits per L-value leads to a performance loss of up to $1$ dB.
	
	\subsection{Generalization Performance}
	
\begin{figure}[t]
		\centering
		\includegraphics[width=3.5in]{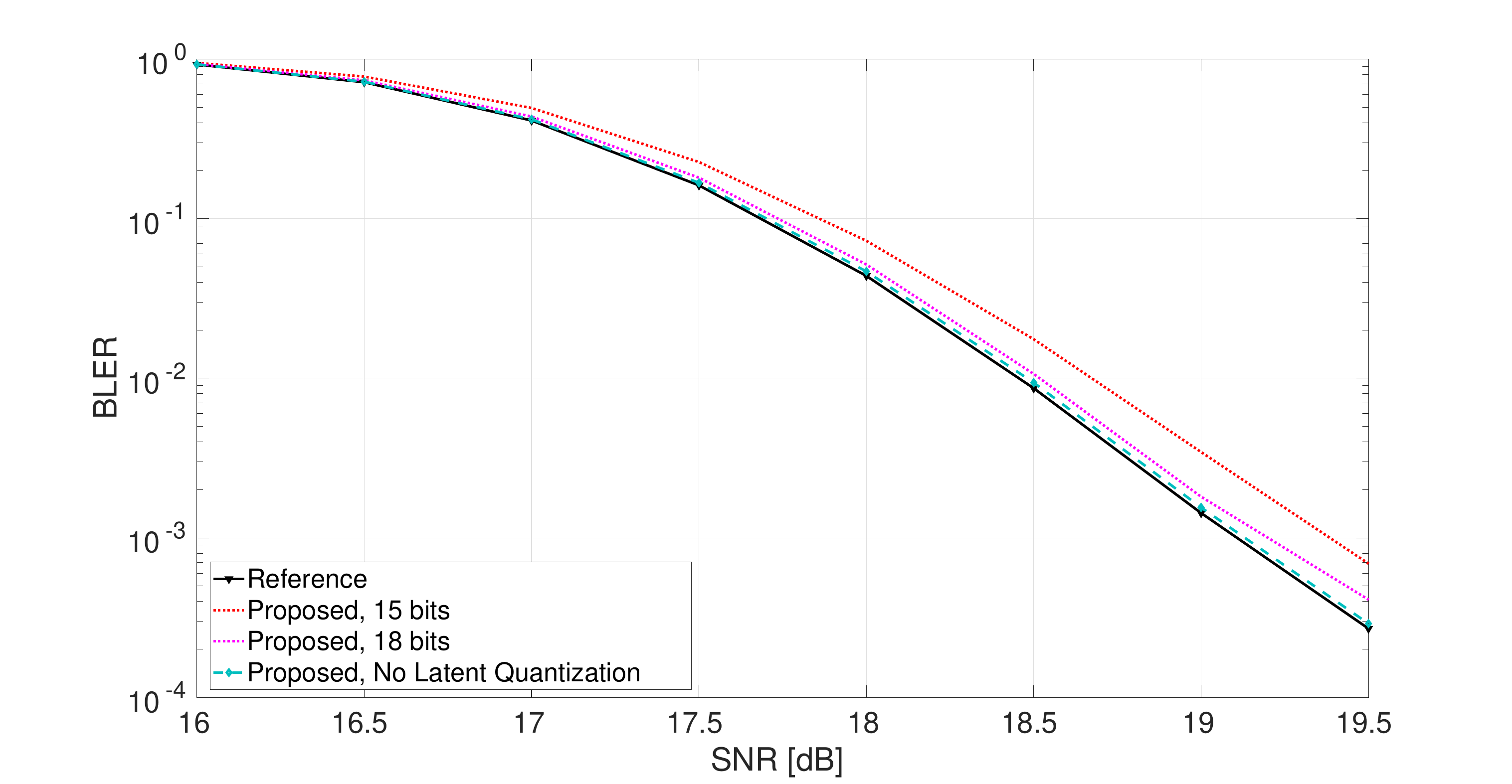}
		\caption{BLER performance of the proposed quantization scheme in a ETU fading scenario with LDPC coding and $K=8$ ($256$-QAM) modulation. The same network and codebooks used in \figurename{ \ref{fig:ldpc_rayleigh}} are reused without additional training or adjustments.}
		\label{fig:ldpc_fading}
	\end{figure}

	We investigate the performance of the proposed quantization scheme when applied to different testing configurations than the one we trained the network with. \figurename{ \ref{fig:ldpc_fading}} shows the validation performance of the network trained in Section \ref{subsect_base_results} when the channel is a frequency domain representation of a multitap fading extended typical urban (ETU) channel, corresponding to an OFDM scenario with a carrier frequency of $2$ GHz, bandwidth $10$ MHz and $81$ subcarriers (one for each QAM symbol). Each LDPC codeword experiences a different realization of the ETU channel and we use a random, but fixed interleaver for extra robustness. The relative gain of our method versus the maximum mutual information quantizer is fully preserved both for $15$ and $18$ bits, even though the network and codebooks are not further trained or adjusted in any way for this particular scenario.
	
\begin{figure}[t]
		\centering
		\includegraphics[width=3.5in]{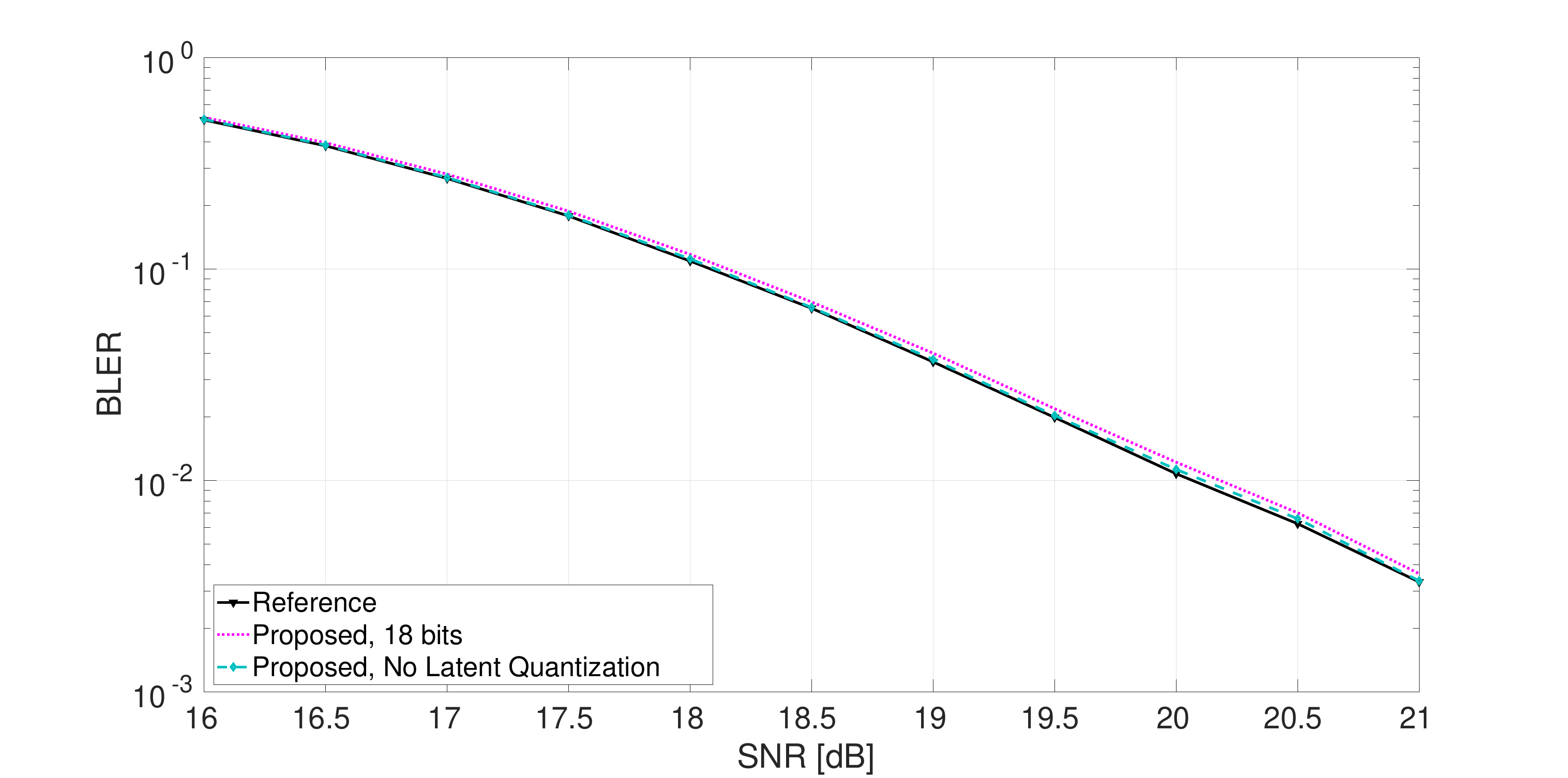}
		\caption{BLER performance of the proposed quantization scheme in a Rayleigh fading scenario with Polar coding and $K=8$ ($256$-QAM) modulation. The same network and codebooks used in \figurename{ \ref{fig:ldpc_rayleigh}} are reused without additional training or adjustments.}
		\label{fig:polar_rayleigh}
	\end{figure}

	Additionally, we investigate the performance of the same network when applied to a Polar-coded Rayleigh fading scenario. We simulate a Polar code of length $256$ and rate $1/2$ used for the New Radio (NR) control channels, decoded with the successive list cancellation algorithm with a list size of $L=4$. \figurename{ \ref{fig:polar_rayleigh}} plots the BLER performance on $\numprint{100000}$ codewords obtained with the same network, where we notice that our method retains its advantages. This leads us to the claim that the learned quantization scheme is universal, in the sense that it exhibits the same performance regardless of the type of channel or channel code used and does not require any further training whatsoever.
	
	\subsection{Discussion}
	The performance of our scheme can be further enhanced if we take into account that it is not required to have the bit resolutions of all latent space components equal when performing quantization. Indeed, judging by \figurename{ \ref{fig:kmeans}} it appears that the second component $z_2$ is more sensitive to scalar quantization since its distribution has a higher entropy. For example, given a budget of $16$ bits, we expect that allocating them to $(z_1, z_2, z_3)$ as $(5, 6, 5)$ instead of, say, $(5, 5, 6)$ will lead to less overall degradation of the end-to-end performance. Because of space constraints and to not make the results too crowded we omit this result, but this is indeed the case. This reasoning can be further extended to subsets of latent components if vector quantization is applied in the latent space, but we leave this for future research.
	
	\section{Conclusions}
	\label{section_conc}
	
	In this work we have introduced a universal log-likelihood ratio (L-value) compression and quantization method that uses a deep autoencoder with a branched decoder, quantizes and reconstructs the latent representation of the set of L-values corresponding to a channel use. The branched decoder architecture allows us to more accurately reconstruct low-magnitude L-values, which are critical for successful decoding under greedy quantization.
	
	Our results show that we can afford quantization with less than two effective bits per L-value for 256-QAM modulation, regardless of the type of channel model or code used with a loss smaller than $0.2$ dB in terms of BLER. In the high performance regime, we achieve losses smaller than $0.05$ dB with an effective $2.25$ bits per L-value. The algorithm can be used for any modulation scheme (with better gains achieved for higher order schemes) and is competitive with state-of-the-art maximum mutual information quantization algorithms, achieving a compression factor of up to two times for the same accuracy.
	
	\bibliographystyle{IEEEtran}
	\bibliography{myBib}
	
\end{document}